\documentclass[letterpaper, 10pt, conference]{ieeeconf}      

\IEEEoverridecommandlockouts                              
\usepackage{graphics} 
\usepackage{graphicx, subfigure}
\usepackage{epsfig} 
\usepackage{mathptmx} 

\usepackage{amsmath}
\usepackage{amsfonts}
\usepackage{amssymb}
\usepackage{algorithm}
\usepackage{algorithmic}
\usepackage[T1]{fontenc}
\usepackage{color}
\usepackage{nomencl}
\usepackage[table]{xcolor}
\usepackage{algorithm}
\usepackage{listings}
\usepackage{float}
\floatstyle{ruled}

\definecolor{propose}{rgb}{0.0,0.6,0.0}
\definecolor{warn}{rgb}{0.7,0.0,0.0}
\definecolor{gray}{rgb}{0.4,0.4,0.4}
\definecolor{darkblue}{rgb}{0.0,0.0,0.6}
\definecolor{cyan}{rgb}{0.0,0.6,0.6}
\definecolor{darkred}{rgb}{0.6,0.0,0.0}

\lstdefinelanguage{XML}
{
  morestring=[b]",
  morestring=[s]{>}{<},
  morecomment=[s]{<?}{?>},
  stringstyle=\color{black},
  identifierstyle=\color{darkblue},
  keywordstyle=\color{cyan},
  morekeywords={xmlns,version}
}

\floatstyle{ruled}
\newfloat{mylisting}{htbp}{loa}
\floatname{mylisting}{Listing}

\title{\LARGE \bf
A Representation Of Robotic Behaviors Using Component Port Arbitration
}

\author{Ali Paikan, Giorgio Metta and Lorenzo Natale
\thanks{The work in this paper was supported by the FP7 EU project Xperience(No. 270273).}
\thanks{A. Paikan, L. Natale and G. Metta are with Italian Institute of Technology (IIT), Genova, Italy. Emails: {\tt\small $\lbrace ali.paikan, lorenzo.natale, giorgio.metta\rbrace @iit.it$.}}%
}

\begin{document}

\maketitle
\thispagestyle{empty}
\pagestyle{empty}

\begin{abstract}
Developing applications considering reactiveness, scalability and re--usability has always been at the center of attention of robotic researchers. Behavior--based architectures have been proposed as a programming paradigm to develop robust and complex behaviors as integration of simpler modules whose activities are directly modulated by sensory feedback or input from other models. The design of behavior based systems, however, becomes increasingly difficult as the complexity of the application grows. This article proposes an approach for modeling and coordinating behaviors in distributed architectures based on port arbitration which clearly separates representation of the behaviors from the composition of the software components. Therefore, based on different behavioral descriptions, the same software components can be reused to implement different applications. 
\end{abstract}

\section{Introduction}
Behavior--based systems (BBSs) have been devised to program robot applications that do not rely on models of the environment and for which reaction to sensory feedback is crucial. However, BBSs are difficult to design when the task involves interaction of large number of software components. Perhaps, this is one of the reasons why behavior--based approaches are not widely applied to the complex robotic applications. The crucial problem is to represent behaviors and components separately so that the latter can be reused more freely. Best practices in robotics~\cite{Brugali2009} promotes the idea that composition and coordination of software component should be separated during the software development life--cycle. Moreover, a proper abstract representation of the behaviors is crucial for the development of complex robotic application. In modern robotic middlewares, coordination of software modules is more difficult since components run asynchronously and are distributed across a network of computers. 
 
\begin{figure}[t]
  \begin{center}
    \includegraphics[width=2.4in]{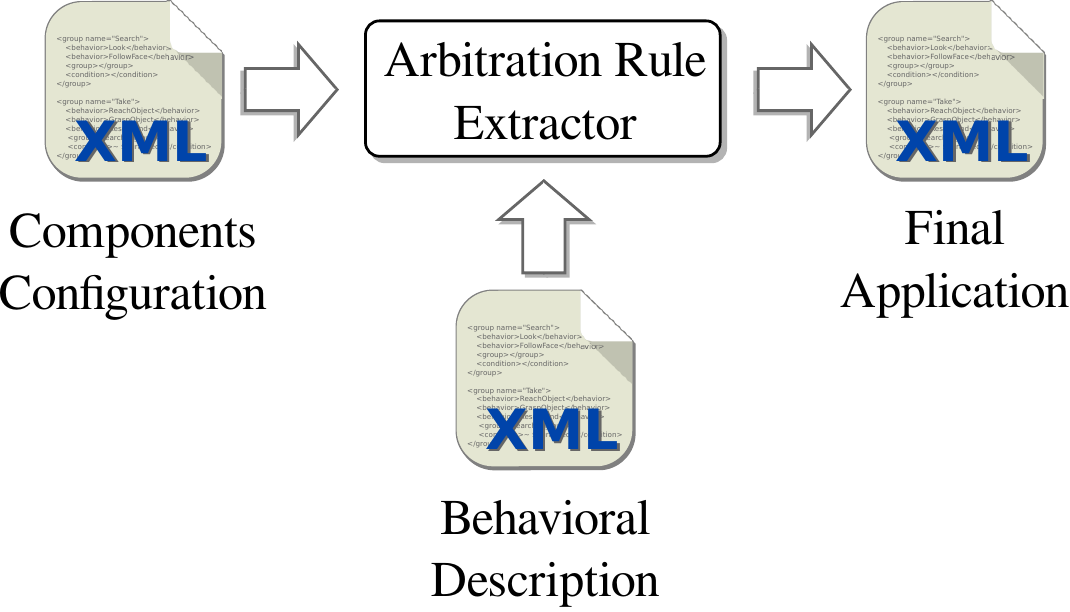}
    \caption{Application generation from behavioral description.}
    \label{fig4:model2appp}
  \end{center}
\end{figure}

We introduce a mechanism for modeling and coordination of behaviors based on port arbitration~\cite{paikan13}. Coordination between modules is achieved by defining a set of rules that specify how to arbitrate conflicts between modules that run concurrently and compete for the same resources. In our approach, building an application out of reusable software is done in two phases (Figure~\ref{fig4:model2appp}). First, software components are configured and interconnected in the (distributed) system. Second, a behavioral model is developed which describes the desired behavior of the system. Coordination is then defined by extracting a set of rules from the behavioral model. These rules determine how data is allowed to travel across the network of components and therefore implicitly define which components are inhibited or free to run. Based on different behavioral description, the same software components can be reused to implement different robotic applications. 

\section{Related work}
Our proposed idea has some similarities with the coordination models and languages which are originated from the work of Gelernter and Carriero~\cite{Gel92}. A coordination model provides a framework in which the interaction of active and independent entities can be expressed using a specific language. An example of coordination languages is Linda~\cite{gelernter1985generative}. This language defines a mechanism to coordinate concurrent computations by means of messages which are formated in tuple structure and can be added to the computation environment. It has been argued that coordination models similar to Linda \footnote{also known as endogenous languages} have the fundamental drawback of intermixing computation with coordination~\cite{Arbab98}. In contrast to Linda which requires computations to make use of specific primitives for coordination, Reo~\cite{Arbab2004} provides a paradigm for composition of distributed software components and services based on the notion of connectors. Reo enforces an exogenous channel--based coordination model that defines how designers can build connectors, out of simpler ones. Application designers can use Reo for compositional construction of connectors that coordinate the cooperative behavior of components in a component--based system. A comprehensive and detailed survey of coordination languages is also given in~\cite{Arbab98}. 

Despite coordination languages have had a large development in the context of parallel and distributed system during the last two decades, to our knowledge, there have not been enough workaround for their realization over available robotic frameworks or study of the required component models for their implementation. In contrast, our proposed coordination model exploits component data--flow ports and the connections among the components which commonly can be found in every nowadays robotic frameworks.

\section{Port-Arbitrated Coordination}
\label{arbitration}
The idea of the port arbitration and its applications in robotics has already been discussed in our previous works~\cite{paikan13}, \cite{paikan2014-portmonitor} and \cite{paikan2014-reusability}. However, in the following, we shortly describe the concept of component coordination using port arbitration mechanism. 

A simple software component can be seen as a computational unit with a set of data--flow ports for communication. It has a set of input ports to receive information from other components and a set of output ports to stream out the data. A component checks for the condition in which to become active (e.g. upon receiving data), processes and sends the results through its output ports. We make the following assumptions for each component:%

\begin{itemize}
	\item The preconditions in which a component gets activated are local to the component itself and they are not visible to other modules. In other words components cannot directly activate or deactivate others. 
	\item Data is streamed out if and only if the component is active. For example, an object detector sends object position information through its output port only if the object has been detected. 
\end{itemize} 

We focus on the typical scenario of a publish--subscribe architecture. The output of a component can be connected to one or more input ports of other component. Interestingly it is also possible to connect multiple outputs to the same input of a component. In the example from Figure \ref{fig:connection}, Object Detector is a component which processes the streamed images from the robot camera and produces the 3D position of the object when detected. Its output is connected to Gaze Control which receives a 3D position in the robot root frame and controls the head of the robot to gaze at the target point. The output port of Object Detector is also connected to the Arm Control component which moves the robot arm's end-effector to reach the target point received from its input port. A crucial aspect is that multiple outputs can be connected to the input port of a component. Without proper coordination among the components, data from different components can be delivered to an input port at any time, potentially causing conflicts. For example, the outputs of Object Detector and Rest Arm are both connected to the input port of Arm Control. These components conflict when they are both active and concurrently send position data to Arm Control. To solve this problem, every input port has an arbitrator which can be configured with a set of rules to properly arbitrate the data received from multiple sources. We propose a mechanism to describe the behavioral model of the task. This model allows to derive the necessary rules that properly configure the arbitrators to implement the task. We define the ingredients of our port-arbitrated coordination system as follow:   
\begin{figure}[t]
  \begin{center}
    \includegraphics[width=3.2in]{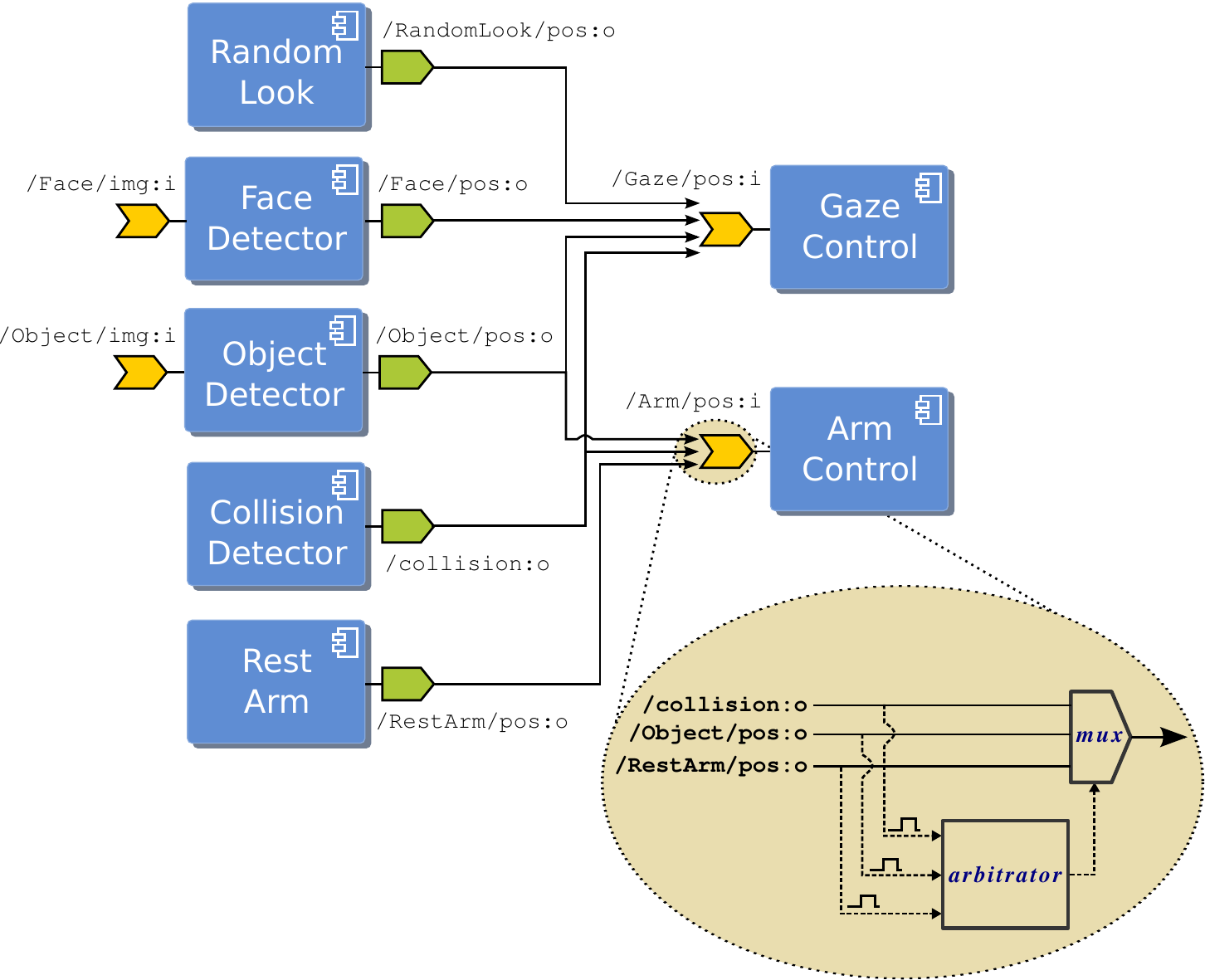}
    \caption{An example of different components and the connections among them. The shaded box represents how an arbitrator is employed in the input port of a component to select between multiple connections to the same port.}
    \label{fig:connection}
  \end{center}
\end{figure}

\begin{itemize}		
    \item A pair of source and destination names which identifies the \emph{connection} from an input port to an output port (e.g. \{\texttt{/Face/pos:o}, \texttt{/Gaze/pos:i}\}).
	
    \item An \emph{active} connection is a connection which has recently delivered data. When data arrives to an input port from a connection, the latter becomes active and remains active for a constant time $T$. The connection will be inactive if no more data arrives within time $T$. Notice that the activation of a connection is defined solely in terms of the data it delivers to the port and irrespectively of the result of the arbitration.
\end{itemize}

\subsection{What are the rules and how they are used in arbitrator?}
Each input port has an arbitrator which selects a single connection among the ones that are active at each time. This ``winner'' connection delivers data to the component whereas data from the other connections gets discarded. This concept is drawn in Figure \ref{fig:connection} for the case of three connections. As it is shown in the figure (shaded part), the port arbitrator is implemented as a multiplexer that let, at most, one active connection deliver its data to the component at each time. In Figure \ref{fig:connection}, Rest Arm periodically sends the resting position of the arm through \texttt{RestArm/pos:o} to Arm Control. This causes the robot to park and keep the arm in the resting position. To grasp an object we want to hand over control of the arm to another component, Object Detector, that sends the position of the object to be grasped to Arm Control. This can be done by inhibiting the connection from \texttt{/RestArm/pos:o} to \texttt{/Arm/pos:i} in the arbitrator of \texttt{/Arm/pos:i}. In other words we want to specify a rule so that the connection \{\texttt{/RestArm/pos:o, /Arm/pos:i}\} can be selected only if connection \{\texttt{/Object/pos:o, /Arm/pos:i}\} is inactive, formally:
\begin{align*}
\mathtt{/RestArm/pos:o \; \land \; \neg /Object/pos:o} \\
\mathtt{\Rightarrow \; Select(/RestArm/pos:o)}
\end{align*}

Suppose now we add another component which is responsible for stopping the the arm upon collision. This component is called Collision Detector in Figure~\ref{fig:connection} and it sends status messages through the port \texttt{/collision:o} when it detects that the arm collides with an object (e.g. using tactile or torque sensors). The desired behavior can be achieved by adding rules in \texttt{/Arm/pos:i} so that activation of \texttt{/RestArm/pos:o} and \texttt{/Object/pos:o} is inhibited by \texttt{/collision:o}, i.e.:%
\begin{equation*}
   \begin{array}[b]{r}
\mathtt{/RestArm/pos:o \; \land \; \neg /Object/pos:o}\\
\mathtt{ \; \land \; \neg /collision:o \Rightarrow \; Select(/RestArm/pos:o)}\\
\\
\mathtt{/Object/pos:o \; \land \; \neg /collision:o} \\
\mathtt{\Rightarrow \; Select(/Object/pos:o)}
    \end{array}
\end{equation*}

Notice that if no rules specifically select a connection, the latter can never deliver data to the component. In our example, no rule is written for \texttt{/collision:o} in Arm Control; so Collision Detector will never deliver data to this component. As it is, its activation state is only used in the evaluation of the rules of other connections. 

When data arrives from a connection to an input port, the corresponding arbitrator has to decide whether to accept or discard it. First the port updates the state of each connection (i.e. active or inactive). Based on the activation states, the required Boolean inputs for the arbitrator are generated (i.e. True for the active and False for the inactive connections). The rules are structured in Binary Decision Diagrams (BDD)~\cite{bryant1986graph} in the arbitrator. By reasoning on the BDD, the arbitrator evaluates whether the connection discards or delivers data to the component.


\section{Modeling the behaviors}
In general, behavior-based robotic controllers consist in a collection of behaviors and are implemented as control laws to achieve and/or maintain goals~\cite{Mataric}. In our approach, behaviors can be described as a set of rules for the port arbitrators. In Figure~\ref{fig:connection}, Face Detector sends the position of the detected face to Gaze Control to follow it. In other words, to implement a behavior called \emph{Follow Face}, the connection \texttt{/Face/pos:o} to \texttt{/Gaze/pos:i} should exist and be selected by the port arbitrator in \texttt{/Gaze/pos:i}. To implement another behavior we call \emph{Look Around}, the connection \{\texttt{/RandomLook/pos:o, /Gaze/pos:i}\} should be selected by the port arbitrator in \texttt{/Gaze/pos:i} to deliver the random position data generated by Random Look to the Gaze Control. Desired behaviors, therefore, can be implemented by selecting connections which are required to deliver data among specific components. At the behavioral description level, we concentrate only on the connections among the components and the necessary rules that select these connections in the arbitrators. The rules can be provided by specifying \emph{configuration} of the connections required for implementing the behavior, under which \emph{condition} the behavior can be activated and the list of behaviors it should \emph{inhibit}.           

\label{modeling}
\begin{figure}[t]
  \begin{center}
    \includegraphics[width=3.0in]{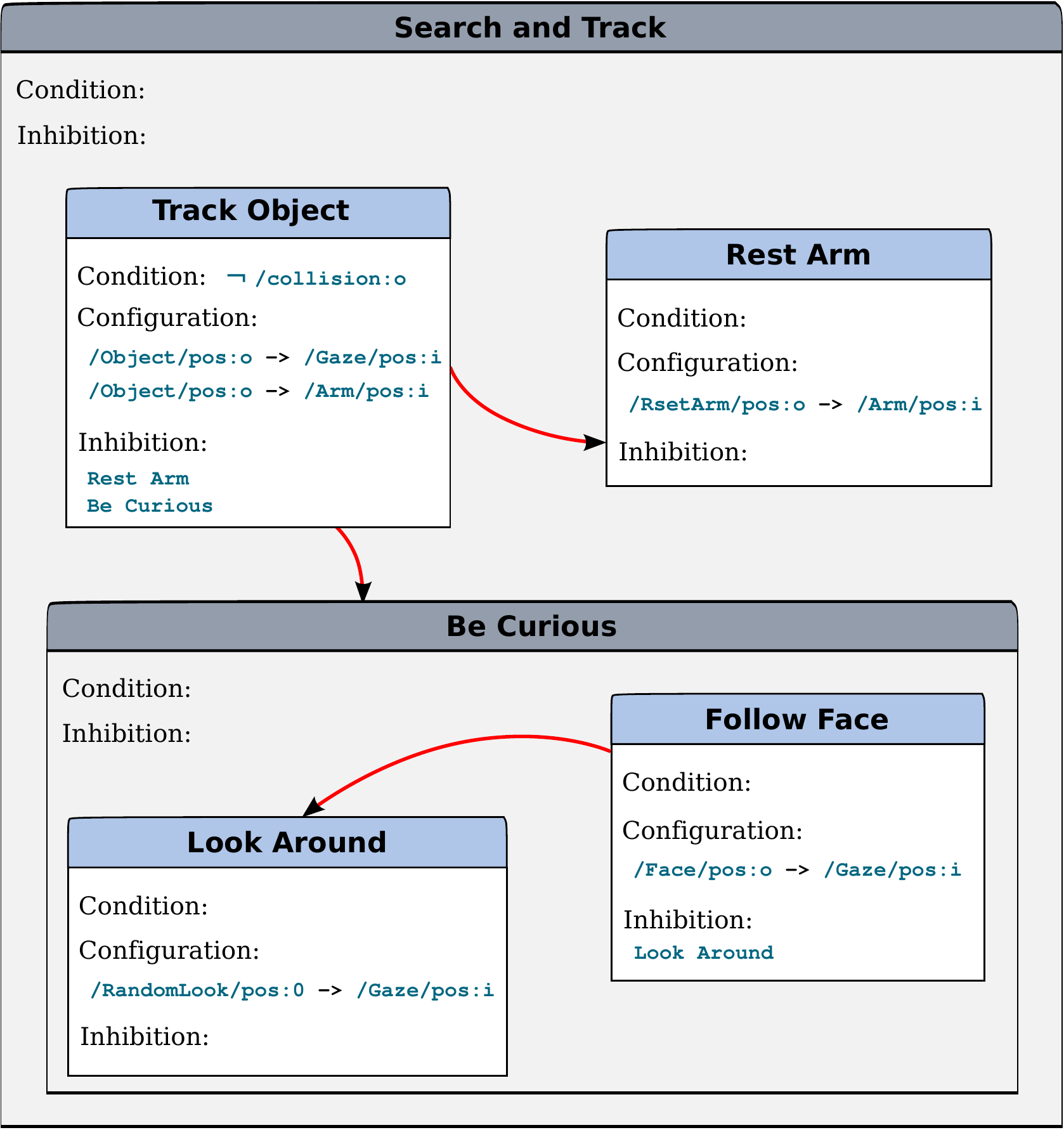}
    \caption{An example of the behavioral model that uses the components from Figure~\ref{fig:connection} to implement the behavior Search and Track. This behavior allows the robot to look around in search for a face or an object. When the robot detects a face it tracks it with the gaze. When it detects an objects it follows it with the gaze and reaches for it. The overall behavior is implemented by coordinating simpler behaviors. Correct coordination is implemented by inhibitions among behaviors (red arrows). E.g., the red arrow from Follow Face to Look around gives higher priority to the first behavior whose activation inhibits
    the second. See also the description in the text.}
    \label{fig:behavior}
  \end{center}
\end{figure}

In Figure~\ref{fig:connection} we have shown an example of the composition of some components and their connections.  Based on them, in Figure~\ref{fig:behavior} we depict an example of behavioral description for a task in which the robot searches around, follows human faces and tracks an object with the hand. In the figure, \emph{Follow Face, Look Around, Rest Arm} and \emph{Track Object} represent behaviors. \emph{Follow Face} and \emph{Look Around} are grouped together to describe, by composition, another (meta behavior) behavior called \emph{Be Curious}. \emph{Be Curious} implements a behavior that let the robot randomly look around or follow a human face if a person appears in the scene. The red arrow from \emph{Follow Face} to \emph{Look around} gives higher priority to the first behavior whose activation inhibits the second. \emph{Rest Arm} describes a behavior that keeps the arm in the resting position. \emph{Track Object} implements tracking of an object with the gaze and reaching for it with the hand. It also inhibits \emph{Rest Arm} and the meta behavior \emph{Be Curious} to prevent them from interfering during tracking. The condition ``$\mathtt{\neg~/collision}$'' in \emph{Track Object} implies that robot can track an object only in absence of collisions. \emph{Track Object}, \emph{Rest Arm} and \emph{Be curious} are further grouped to describe another meta behavior we called \emph{Search and Track}. 
    
\subsection{Behavior Specification}
A behavior (or a meta behavior) has the following properties: 

\emph{Configuration} of a behavior is the list of connections which should be selected by the port arbitrators to implement the behavior. For meta-behaviors, configuration is as a list of behaviors or other meta-behaviors. For example, in Figure~\ref{fig:behavior}, the configuration property of \textit{Track Object} implies that to follow an object with the head, \texttt{/Object/pos:o} should feed data to Gaze Control at \texttt{/Gaze/pos:i}. Tracking with the hand is achieved by sending \texttt{/Object/pos:o} to Arm Control  at \texttt{/Arm/pos:i}. Notice that here we focus only on the connections which define the behavior of the system, but other connections are required for proper functioning of some modules (e.g. Face Detector and Object Detector require connections from the robot cameras) For simplicity we do not consider these connections here.

\emph{Condition} is an optional property which specifies in, first--order logic, a constraint that should be verified for the behavior to be activated. The condition $\mathtt{\neg~/collision:o}$ of \textit{Track Object} requires that all the connections specified in its Configuration should be selected only if the port \texttt{/collision:o} is inactive (i.e. it is not sending messages). In a meta-behavior the Condition affects all its child behaviors, i.e. conditions from all parent meta-behaviors in a hierarchy are conjuncted and inherited by all child behaviors.

\emph{Inhibition}, specifies inhibitions between behaviors or meta-behaviors. Specifying inhibitions allows coordinating behaviors that are competing for the same resources. In Figure~\ref{fig:behavior} we define the behavior \textit{Look Around} which is implemented by connecting ports of Random Look to Gaze Control. We also define \textit{Follow Face} and \textit{Track Object}. These behaviors compete to control the gaze of the robot by sending commands to Gaze Control at \texttt{/Gaze/pos:i}. Conflicts are avoided by further specifying the overall behavior of the robot and assigning inhibitions. In Figure~\ref{fig:behavior} \textit{Follow Face} inhibits \textit{Look Around}. In more details this tells the arbitrator in \texttt{/Gaze/pos:i} that connection \{\texttt{/RandomLook/pos:o, /Gaze/pos:i}\} should not be selected when connection \{\texttt{/Face/pos:o, /Gaze/pos:i}\} is active, because Face Detector is sending data to Gaze Control. 
A behavior can also inhibits a meta-behavior. In this case, the behavior inhibits all the behaviors in the meta-behavior. In Figure~\ref{fig:behavior}, \emph{Track Object} inhibits \emph{Be Curious}, i.e. it inhibits \emph{Follow Face} and \emph{Look around}. In practice this corresponds to assigning decreasing priorities to \emph{Track Object}, \emph{Follow Face} and \emph{Look around} to avoid conflicts in Gaze Control. Similar rules are applied if a meta-behavior inhibits another behavior or another meta-behavior. For the sake of modularity and reusability, behaviors can only inhibit other behaviors within the same meta-behavior. For example in Figure~\ref{fig:behavior}, \emph{Follow Face} is not allowed to inhibit \emph{Rest Arm}. 

\section{Extracting Rules from Behavioral Model }
\label{rules}
In the previous section we have described how behaviors are modeled using connections between ports. In this section we explain how the necessary rules for the arbitrators are extracted from the behavioral model. Every behavior has a list of connections specified by its Configuration property. The properties Conditions and Inhibition determine an extra set of constraints that are applied to the port arbitrators of its connections. 

For example in Figure~\ref{fig:behavior}, \emph{Look Around} is inhibited by \emph{Follow Face}. Both behaviors have a connection to \texttt{/Gaze/pos:i}; Thus the following rule is added to the port arbitrator in \texttt{/Gaze/pos:i}: 
\begin{align*}
\mathtt{/RandomLook/pos:o \; \land \; \neg /Face/pos:o} \\
\mathtt{\Rightarrow \; Select(/RandomLook/pos:o)}
\end{align*}

\emph{Look Around} is also inhibited by \emph{Track Object} (through the inhibition to \emph{Be Curious}). Therefore the previous rule is updated with ``$\mathtt{\neg /Object/pos:o}$'' to reflect the new constraint: 
\begin{align*}
\mathtt{/RandomLook/pos:o \; \land \; \neg /Face/pos:o} \\
\mathtt{\land \; \neg /Object/pos:o \;\; \Rightarrow \; Select(/RandomLook/pos:o)}
\end{align*}

A behavior's Condition and the conditions that are inherited from the parent groups are also added in the same way to the port arbitrators of all of the connections specified in Configuration. For example, the constraint ``$\mathtt{\neg /collision:o}$'' is added to the rules for \texttt{/Object/pos:o} in the arbitrators at \texttt{/Gaze/pos:i} and \texttt{/Arm/pos:i}. 

To summarize, the algorithm to extract the arbitration rules from the behavior model is easily done in two steps for each behavior \emph{i} in the model. First: the Condition of \emph{i} is updated with all the conditions it inherits from the parent meta-behaviors. This condition is added as an extra constraint to the port arbitrators of all the connections specified in Configuration. Second: further conditions are extracted from all inhibitors of \emph{i} and added to the rules of the corresponding port arbitrators.\\

\begin{minipage}{\linewidth}   
\lstset{%
label=lst4:behmodel,
caption={The representation of "Be Curious" behavior in XML format.},
frame=single,
captionpos=b,
numbers=none,
framexleftmargin=3pt,
framexrightmargin=-5pt
}
\begin{lstlisting}[language=XML]
<define name="gaze"> /Gaze/pos:i </define>

<meta_behavior name="Be Curious">
   <behavior>Look Around</behavior>
   <behavior>Follow Face</behavior>
   <condition></condition>
   <inhibition></inhibition>
</meta_behavior>

<behavior name="Look Around">
   <config at="${gaze}">/RandomLook/pos:o</config>
   <condition></condition>
   <inhibition></inhibition>
</behavior>

<behavior name="Follow Face">
   <config at="${gaze}">/Face/pos:o</config>
   <condition></condition>
   <inhibition>Look Around</inhibition>
</behavior>
\end{lstlisting} 
\end{minipage}

The behavioral model can be represented using Extensible Markup Language (XML). Listings~\ref{lst4:behmodel} illustrates the representation of \emph{Be Curious} behavior in the XML format. The model are used by a third--party tool from the YARP~\cite{yarp06} robotic framework to extract the arbitration rules and update the configuration of connections. The concept is also illustrated in Figure~\ref{fig4:model2appp}. YARP offers a method to describe the configuration of components and their connections in XML format which known as application description file. In short, an application description file contains all the required modules, their configuration (i.e., parameters), the way they are interconnected, and the necessary information for their deployment. The required information for orchestration (coordination) of these modules can be represented in another XML file (similar to Listing~\ref{lst4:behmodel}) using our behavioral model based on port--arbitrated mechanism. The final application, can be generated by extracting the arbitration rules from the behavioral description file and using them to configure the component connections from the application description file. Therefore, based on different behavioral model, the same components the their configuration can be used for the development of different robotic applications. 

\section{Discussions}
Different behavior selection mechanisms are compared in~\cite{pirjanian1999behavior, MacKenzieDouglasCandArkinRonaldCandCameron1997} and~\cite{scheutz2004architectural}. An alternative approach to competitive action selection is a cooperative mechanism in which recommendations from multiple behaviors are combined to form a control action that represents their consensus. An example of this type of mechanism is DAMN~\cite{Rosenblatt1997}. It uses a centralized arbitrator to fuse the collected commands from different behaviors and select the action which best satisfies the prioritized goals of the system. Nowadays, due to heterogeneity of data types and the complexity of the control systems, the proposed methodology is practically limited to low-level control. However, in our approach, behaviors can also help each other to become active by enabling the relevant connections. This has already been discussed in detail in~\cite{paikan13}.

Even though coordination based on port arbitration can cover a wide variety of robotic applications, we have experienced certain limitations in the system. First, since arbitration is usually done on the data from unidirectional connection to an input port, it cannot be easily used in a service--oriented system where interactions between modules are bidirectional and done using blocking remote procedure
calls. Moreover, a robotic task might require performing a sequence of actions synchronized with the internal state of components. This can be also made using port arbitration, nevertheless, delegating this responsibility to a dedicated, external component can be preferable in favor of simplicity and performance.


\section{Conclusion}
\label{conclusion}
This work--in--progress article has introduced a mechanism based on port arbitration for modeling and coordination of robotic behaviors. We have shown how robotics tasks can be represented using our behavioral description model and coordinated in a distributed component--based framework using port--arbitrated mechanism and without using any central coordinator. Remarkably, We demonstrated that in our framework, based on different behavioral descriptions, several robotic applications can be hierarchically implemented using the same reusable software components.   
 
Future works will involve investigating proper methods for the validation of the model and checking the consistency of the rules which are extracted from the behavioral description during the development of the tasks.

\section*{Acknowledgments}
The research leading to these results has received funding from the European Community's Seventh Framework Programme (FP7 ICT) under grant agreement No. 270273 (Xperience) and 611832 (WALK-MAN).

\bibliographystyle{IEEEtran}
\bibliography{references}

\end{document}